\documentclass[runningheads]{llncs}

\usepackage[T1]{fontenc}

\usepackage{graphicx}%
\usepackage{multirow}%
\usepackage{amsmath,amssymb,amsfonts}%
\usepackage{mathrsfs}%
\usepackage[title]{appendix}%
\usepackage{xcolor}%
\usepackage{textcomp}%
\usepackage{manyfoot}%
\usepackage{booktabs}%
\usepackage{algorithm}%
\usepackage{algorithmicx}%
\usepackage{algpseudocode}%
\usepackage{listings}%
\usepackage{xspace}%
\usepackage{makecell}
\usepackage{adjustbox}
\usepackage{numprint}
\usepackage{bbding}
\usepackage{subcaption}
\usepackage{relsize}
\usepackage{tikz}
\usepackage{gensymb}

\begin{document}

\newcommand{\mnamens}{DVP} 
\newcommand{\mname}{\mnamens\xspace}

\newcommand{\findent}{\hspace*{\parindent}}

\newcommand{\subfrozen}{\textsubscript{{\relscale{0.9}\SnowflakeChevron}}\xspace}
\newcommand{\subsmall}{\textsubscript{small}\xspace}

\title{Disentangling Visual Priors: Unsupervised Learning of Scene Interpretations with Compositional Autoencoder} 


\titlerunning{Disentangling Visual Priors}

\author{Krzysztof Krawiec\inst{1}\orcidID{0000-0001-5439-3231} 
\\
Antoni Nowinowski\inst{1}\orcidID{0009-0003-6357-8918} }

\authorrunning{K. Krawiec and A. Nowinowski}

\institute{Institute of Computing Science, Poznan University of Technology, Poznań, Poland
\email{antoni.nowinowski@doctorate.put.poznan.pl}}

\maketitle

\begin{abstract}
Contemporary deep learning architectures lack principled means for capturing and handling fundamental visual concepts, like objects, shapes, geometric transforms, and other higher-level structures. We propose a neurosymbolic architecture that uses a domain-specific language to capture selected priors of image formation, including object shape, appearance, categorization, and geometric transforms. We express template programs in that language and learn their parameterization with features extracted from the scene by a convolutional neural network. When executed, the parameterized program produces geometric primitives which are rendered and assessed for correspondence with the scene content and trained via auto-association with gradient. We confront our approach with a baseline method on a synthetic benchmark and demonstrate its capacity to disentangle selected aspects of the image formation process, learn from small data, correct inference in the presence of noise, and out-of-sample generalization. 

\end{abstract}


\section{Introduction}\label{s:intro}

Computer vision (CV) experiences rapid progress thanks to recent advances in deep learning (DL), which keeps outperforming humans not only on benchmarks but also in demanding real-world applications. Unfortunately, the mainstream DL models still lack the capacity of structural, higher-level interpretation of visual information. Most of their impressive feats concern low-level image processing, where local features and patterns are essential. As a result, DL excels at tasks like denoising, superresolution, style transfer, object detection, and similar. However, as soon as higher-level, structural descriptions become essential, DL models tend to overfit, fail, and -- in the generative setting -- hallucinate, producing images that are obviously incoherent for humans. 

The remedy that is nowadays being proposed to address these issues is to throw more data at the model. While this may indeed bring measurable improvements in some cases, the law of diminishing returns renders this strategy impractical: gaining even a tiny improvement may require vast amounts of data, which often needs to be subject to costly labeling. Moreover, empirical and theoretical works indicate that this policy is fundamentally flawed and will never address the problem of the long tails of distributions and out-of-sample generalization (see, e.g., \cite{Cremer_2021}). Recent works suggest that these limitations affect even the most sophisticated and largest architectures, large language models \cite{Wu_2023}. 
It becomes thus evident that contemporary DL still lacks the means for principled capturing and handling of fundamental CV concepts, like objects, shapes, spatial relationships between them, and other higher-level structures. 

In this study, we posit that \emph{the promising avenue for making CV systems capable of principled, scalable scene interpretation is to equip them with elements of domain-specific knowledge about the image formation process}. We achieve that by designing a neurosymbolic architecture of Disentangling Visual Priors (\mname) that uses a domain-specific language (DSL) to capture selected priors of image formation, including object compactness, shape, appearance, categorization and geometric transforms. Using this DSL, we can express template programs that explain scene content and learn how to parameterize them using features extracted from the scene by a convolutional neural network (CNN). The execution of the parameterized program produces geometric primitives that are then rendered and assessed for their correspondence with the scene content. As the DSL programs and scene rendering are realized in a differentiable way, \mname forms a \emph{compositional autoencoder} that is trainable with gradient end-to-end. Crucially, it learns from raw visual data, without object labels or other forms of supervision.  

In the experimental part of this study, we demonstrate how \mname manages to autonomously learn to
\emph{disentangle} several aspects of scene content and image formation: object color from its shape, object shape from its geometric transforms, geometric transforms from each other, and object category from the remaining aspects. As a result, we obtain an interpretable, transparent image formation pipeline, which naturally explains the chain of reasoning, can learn from small data, and conveniently lends itself to further inference (e.g. object classification) and out-of-sample generalization. 

This paper is organized as follows. Section \ref{sec:approach} presents \mname. Section \ref{sec:related} discusses the related work. Section \ref{sec:experiment} covers the results of extensive experimenting with \mname and its juxtaposition with several baseline architectures. Section \ref{sec:conclusions} concludes the paper and points to possible future research directions.

\begin{figure}[t]
    \centering
    \includegraphics[width=.8\textwidth]{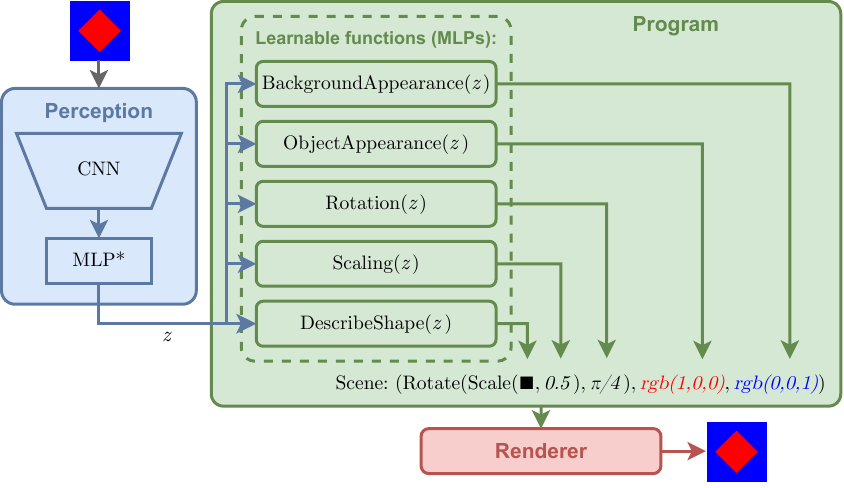}
    \caption{ \mname architecture. Perception encodes an image into a latent vector $z$. Program maps $z$ to a Scene. Renderer renders the Scene as a raster image.  }
    \label{fig:diagram}
\end{figure}

\section{The proposed approach}\label{sec:approach}


In \mname, the model attempts to produce a symbolic interpretation of the perceived scene and learns by trying to align the rendering of that interpretation with the actual scene content. In this study, we consider 2D scenes featuring a single object filled with color and placed at the center of a uniform background. 

\mname is a neurosymbolic architecture comprising three components (Fig. \ref{fig:diagram}):
\begin{enumerate}
    \item The \textbf{Perception} module, a convolutional neural network (CNN) that transforms the 2D input image of the scene into an image latent $z$, 
    \item The DSL \textbf{Program} that generates and transforms visual primitives parameterized with geometric features derived from $z$ and so produces a \emph{hypothesis} (guess) about the scene content in terms of symbolic representation comprising objects and their visual properties,  
    \item The \textbf{Renderer} that renders that symbolic representation on a raster canvas. 
\end{enumerate}
In the following, we describe these modules on the conceptual level; for implementation details and hyperparameter setting, we refer the reader to Sec.\ \ref{sec:experiment}. 

The \textbf{Perception} can be an arbitrary neural network mapping the input image to a fixed-dimensional latent vector $z$, typically a CNN followed by a stack of dense layers (MLP), often referred to as \emph{backbone} in DL jargon. The specific variants of Perception used in this study are covered in Sec.\ \ref{sec:experiment}.

The \textbf{Program} is parameterized with the latent $z$ and expressed in a bespoke DSL that features the following types: 
\begin{itemize}
    \item Latent, the type of $z$, which is the only input to the program, 
    \item Double, for floating-point scalar values,
    \item Appearance, a 3-tuple of Doubles encoding color in the RGB format, 
    \item Shape, which stores a closed silhouette of an object (detailed below),
    \item Scene, a 3-tuple (Shape, Appearance, Appearance) describing respectively the object's shape, its color, and the color of the background. 
\end{itemize}
Programs are composed of \emph{functions}, some of which depend on internal parameters that can change in training. Such \textbf{learnable DSL functions} are:
\begin{itemize}
    \item ObjectAppearance: Latent $\rightarrow$ Appearance 
    \item BackgroundAppearance: Latent $\rightarrow$ Appearance 
    \item Scaling: Latent $\rightarrow$ Double  
    \item Rotation: Latent $\rightarrow$ (Double, Double)
    \item DescribeShape: Latent $\rightarrow$ Shape  
    \item Prototype: Latent $\rightarrow$ Shape 
\end{itemize}
Each learnable function contains a separate dense neural network (MLP) that determines the outcome of function's calculation in direct or indirect manner. The direct mode is used by all but the last function in the above list. For instance, Rotation is an MLP with the number of inputs equal to the dimensionality of $z$ and two outputs that encode the perceived object's predicted rotation angle $\phi$ as $(\cos\phi,\sin\phi)$; Scaling is an MLP that returns a single positive scalar to be interpreted as a magnification factor of the object; DescribeShape contains an MLP that learns to retrieve the shape of the input object observed by Perception and conveyed by the latent $z$, and encode it as a vector of Elliptic Fourier Descriptors (EFD, \cite{kuhl1982}), detailed in \ref{app:efd}. Technically, the output of the MLP is interpreted as real and imaginary parts of complex coefficients that form the spectrum which is mapped with inverse Fourier Transform to object's contour represented as complex `time series' $(x_t,y_t)$.

The Prototype function implements the indirect mode. It holds an array $A$ of shape prototypes, represented as a learnable embedding (i.e. the elements of the array are parameters of this operation). It also contains an MLP which, when queried, returns the predicted index of the prototype. Overall, Prototype returns thus $A[$MLP($z$)$]$; however the indexing operation is implemented in differentiable fashion, which makes it amenable to gradient-based training algorithms. 

The \textbf{non-learnable DSL functions} are:
\begin{itemize}
    \item Scale: (Shape, Double) $\rightarrow$ Shape 
    \item Rotate: (Shape, (Double, Double)) $\rightarrow$ Shape
\end{itemize}
where the former scales the input shape according to the scaling factor given by the second argument, and the latter rotates the input shape by $\phi$ given as  $(\cos\phi,\sin\phi)$. 

The DSL allows expressing a number of programs. For the purpose of this study, we are interested in programs with the signature Latent $\rightarrow$ Scene, where the result is passed to the Renderer. One of the programs used in the experiments in Sec.\ \ref{sec:experiment} is shown in Fig. \ref{fig:diagram} and has the form: 

\vspace{3mm}
$P$($z$: Latent) $\rightarrow$ Scene = (\\ 
\findent\findent Rotate(Scale(DescribeShape($z$), Scaling($z$)), Rotation($z$)), \\
\findent\findent ObjectAppearance($z$), \\
\findent\findent BackgroundAppearance($z$) \\
\findent )
\vspace{3mm}

\noindent where the shape inferred from the latent by DescribeShape is subsequently subject to scaling and rotation, after which the transformed shape is combined with its appearance and the appearance of the background, and returned as a Scene to be rendered. 
In Sec.\ \ref{sec:experiment}, we compare a few variants of \mname equipped with such predefined programs. 

The \textbf{Renderer} is responsible for producing the rasterized and differentiable representation of the scene produced by the Program. In recent years, several approaches to differentiable rendering have been proposed \cite{ravi2020pytorch3d,Li:2020:DVG,KaolinLibrary,Mitsuba3}; of those, we chose the rendering facility available in PyTorch3D \cite{ravi2020pytorch3d} for our implementation of \mname, which was motivated by its versatility and ease of use.\footnote{Pytorch3D can render several classes of geometric objects using representations such as meshes, point clouds, volumetric grids, and neural-network based representations, also in 3D; while this last capability was not used in this study, we plan to engage it in further works on this topic.} This renderer operates similarly to computer graphics pipelines: the scene is approximated with a mesh, the triangles comprising the mesh are rasterized, and the resulting rasters are merged to form the final image. 

\vspace{1mm}\noindent\textbf{Training}.
We train \mname via autoassociation, like conventional autoencoders. As all its components are differentiable, it can be trained with an arbitrary gradient-based algorithm. In each iteration, a training image $x$ (or a batch thereof) is fed into the model, which responds with the raster canvas $\hat x$ = Renderer(Program(Perception($x$))) containing the predicted rendering. Then, a loss function $L$ is applied to these images and measures the pixel-wise error between them. Finally, we take the gradient $\nabla_\theta L(x,\hat x)$ with respect to all parameters of the model $\theta$ and use it to update $\theta$, accordingly to the specific variant of stochastic gradient descent. $\theta$ collects the parameters of Perception and all learnable DSL instructions in the Program. The Renderer is non-parametric; therefore, its only role is to `translate' the gradient produced by $L$ into the updates of $\theta$ required by the Program and the Perception.

\section{Related work}\label{sec:related}

\mname represents the category of image understanding systems inspired by the  \emph{vision as inverse graphics}" \cite{barrow1978recovering}, which can be seen as a CV instance of the broader \emph{analysis-by-synthesis} paradigm. While considered in the CV community for decades (see, e.g., \cite{Krawiec_2007}), it experienced significant advancement in recent years, thanks to the rapid progress of DL that facilitated end-to-end learning of complex, multi-staged architectures. Below, we review selected representatives of this research thread; for a thorough review of other approaches to compositional scene representation via reconstruction, see \cite{Yuan_Chen_Li_Xue_2023}; also, \cite{Elich_2022} contains a compact review of numerous works on related CV topics, including learning object geometry and multi-object scene representations, segmentation, and shape estimation. 

The Multi-Object Network (MONet), proposed in \cite{Burgess_2019} and used in the experimental part of this study as a baseline, is a composite unsupervised architecture that combines image segmentation based on an attention mechanism (to delineate image components) with a variational autoencoder (VAE), for rendering individual components in the scene. As such, the approach does not involve geometric aspects of image formation and scene understanding. Also, it does not involve geometric rendering of objects: the subimages of individual components are generated with the VAE and `inpainted' into the scene using raster masks.  

PriSMONet \cite{Elich_2022} attempts decomposition of 3D scenes based on their 2D views. Similarly to MONet, it parses the scene sequentially, object by object, and learns to generated objects' views composed from several aspects: shape, textural appearance and 3D extrinsics (position, orientation and scale). The background is handled separately. Object shapes are represented using the Signed Distance Function formalism, well known in CV, and generated using a separate autoencoder submodel (DeepSDF); in this sense, PriSMONet does not involve shape priors. In contrast to \cite{Burgess_2019}, the architecture engages differentiable rendering. 

Another related research direction concerns part discovery, where the goal is to decompose the objects comprising the scene into constituents, which preferably should have well-defined semantics (i.e., segmentation at the part level, not the object level). The approaches proposed therein usually rely on mask-based representations (see, e.g. \cite{Hung2019,Choudhury2022}); some of them involve also geometric transforms (e.g. \cite{Hung2019}). 

\mname distinguishes itself from the above-cited works in several ways. Firstly, it relies on a physically plausible, inherently differentiable, low-dimensional shape representation (EFD), while most other works use high-dimensional and localized representations, like pixel masks, point clouds, meshes, or signed distance functions (see \cite{Elich_2022} for review). 
\mname represents geometric transforms explicitly, rather than as an implicit latent, like e.g.  \cite{Burgess_2019}. Last but not least, it expresses the image formation process in a DSL, which facilitates disentanglement of multiple aspects of this process, i.e. object shape, appearance, and pose.

\section{Experimental results}\label{sec:experiment}

We compare the performance of several variants of \mname to related methods and assess its ability to learn from small data and robustness to noise. 

\vspace{1mm}\noindent\textbf{Task formulation}.
One of the most popular benchmarks for compositional scene interpretation is Multi-dSprites \cite{Burgess_2019}. 
In this study, we consider a similar problem but involving a single object. We generated a dataset of 100{,}000 
images, each containing a single shape from one of 3 categories (ellipse, square, heart), randomly scaled and rotated, and rendered using a randomly picked color at the center of a 64x64 raster filled with a different random color. The task of the model is to reproduce this simple scene in a compositional fashion. The dataset was subsequently divided into training, validation, and test subsets of, respectively, 90k, 5k, and 5k examples. While our dataset is similar to dSprites \cite{dsprites17}, it diverges from it in centering objects in the scene, using color and a larger range of object sizes, and applying anti-aliasing in rendering.  

\vspace{1mm}\noindent\textbf{Configurations of \mname and baselines}. 
We compare \mname architectures that feature two types of DSL programs, those based on \emph{direct} inference of the object shape from the latent (\mname-D), and those based on object \emph{prototypes} (\mname-P). In the former, we employ the program $P$ presented in Sec.\ \ref{sec:approach}. In \mname-P, we replace in $P$ the Describe($z$) call with Prototype($z$). 

\begin{table}[t]
\centering
\caption{The configurations of \mname and the baseline models.}
\label{tab:configs}
\begin{tabular}{lcrrc}
\toprule
Name & Perception (backbone) & \multicolumn{2}{c}{Number of parameters} &  Frozen backbone \\ 
\cmidrule(rl){3-4}
& & \makecell[c]{Total} & \makecell[c]{Perception} \\
\midrule
\mname-D\subfrozen & ConvNeXt-B & 107,833,773 & 87,564,416 & yes \\
\mname-P\subfrozen & ConvNeXt-B & 107,827,861 & 87,564,416 & yes \\
\mname-D\subsmall & CNN1 & 7,940,907 & 4,504,320 & no \\
\mname-P\subsmall & CNN1 & 7,928,851 & 4,504,320 & no \\
\midrule
MONet\subfrozen & ConvNeXt-B & 88,598,293 & 87,923,599 & yes \\
MONet & CNN2 & 896,486 & 221,792 & no \\
\bottomrule
\end{tabular}
\end{table}

We consider two categories of Perception modules (`backbones'), i.e. subnetworks that map the raster image to a fixed-dimensional latent vector (Sec.\ \ref{sec:approach}): pre-trained and not pre-trained ones (Table \ref{tab:configs}). Our pre-trained architecture of choice is ConvNeXt-B \cite{Liu_2022}, a large modern model that proved very effective at many computer vision tasks \cite{Goldblum_2023}. 
In the non-pretrained case, Perception is trained from scratch alongside the rest of the model. For this variant, Perception is a 6-layer CNN (CNN1) followed by an MLP (see \ref{app:model} for details).

Our baseline model is MONet \cite{Burgess_2019}, outlined in Sec.\ \ref{sec:related}. To provide for possibly fair comparison, we devise its pre-trained and non-pretrained variant: in the former, we combine it with ConvNeXt-B serving as part of the feature extraction backbone network (the counterpart of Perception in \mname); in the latter it is the original CNN used in MONet (CNN2 in Table \ref{tab:configs}). 

The models were trained using the Adam algorithm \cite{DBLP:journals/corr/KingmaB14} with the learning rate 0.0001. The training lasted for 40 epochs, except for \mnamens\subsmall configurations trained on the full dataset, which were trained for 160 epochs. A typical training run lasted 3 to 4 hours on a PC with NVIDIA GeForce RTX 3090 GPU.  


\vspace{1mm}\noindent\textbf{Data scalability}. 
We expect the compositional constraints imposed by the DSL to narrow the space of possible scene interpretations and facilitate learning from small data, so we trained \mname and the baseline architectures in three scenarios: on the entire training set (100\%, 90k examples) and on the training set reduced (via random sampling) to 5\% and 1\%, i.e. respectively 4.5k and 900 examples. 

In Table \ref{tab:accuracy}, we juxtapose the test-set reconstruction accuracy of \mname with the reference configs using commonly used metrics: Mean Square Error (MSE), Structural Similarity Measure (SSIM, \cite{Wang2004}), Intersection over Union (IoU), and Adjusted Rand Index (ARI\footnote{ARI measures the similarity of two clusterings by counting the numbers of pairs of examples that belong to the same/different clusters in them and adjusting those numbers for the odds of chance agreement. Here, the examples are pixels, and there are two clusters: the background and the foreground.}). 
While MSE is calculated directly from the RGB values of the input image and model's rendering (scaled to the $[0,1]$ interval), IoU and ARI require the rendered pixels to be unambiguously assigned to objects or the background. 
We achieve that by forcing the models to render scenes with white objects on a black background, which results in binary masks representing the assignment of pixels\footnote{Our implementation reuses object masks produced in the RGB rendering process.} (in contrast to complete rendering, where the model controls also the colors). 

When training models on the full training set (100\%), \mname is clearly worse than MONet on MSE, which can be explained by the latter using raster masks to delineate objects from the background. Nevertheless, the remaining metrics suggest that the gap between the methods is not that big; in particular, when using the pre-trained large perception, \mname in the direct mode manages to perform almost on par on SSIM and beats MONet on the IoU. 

When trained on 5\% examples from the original training set, all methods observe deterioration on all metrics (though MONet configurations maintain almost unaffected IoU and ARI); this is particularly evident for the MSE, which increases several folds for all configurations. When training on 1\% of the original training set, all configurations experience further deterioration on all metrics. However, this time MONet seems to be more affected than \mname, in particular on MSE (almost 2 orders of magnitude compared to the 5\% scenario) and on the IoU (over 20 percent point loss for both pre-trained and non-pretrained variant). In contrast, MSE for \mname increases by a single-digit factor, and other metrics drop only moderately. This confirms that \mname is capable of learning effectively from small data, also when forced to train the Perception from scratch.  

\begin{table}[t]
\centering
\caption{Reconstruction accuracy of \mname and the baselines. The best results are highlighted in bold. } 
\label{tab:accuracy}
\begin{tabular}{lcrrrrr} 
\toprule
Method & Training set size & MSE & SSIM & IoU & ARI & \\
\midrule
\mname-D\subfrozen & 100\% & 0.000305 & 0.9864 & \textbf{0.9852} & 0.9843 \\
\mname-P\subfrozen & 100\% & 0.000606 & 0.9745 & 0.9772 & 0.9754 \\
\mname-D\subsmall  & 100\% & 0.001597 & 0.9264 & 0.9115 & 0.9028 \\
\mname-P\subsmall  & 100\% & 0.002636 & 0.8960 & 0.8767 & 0.8679 \\
MONet              & 100\% & 0.000141 & 0.9889 & 0.9754 & \textbf{0.9877} \\
MONet\subfrozen    & 100\% & \textbf{0.000137} & \textbf{0.9897} & 0.9751 & 0.9875 \\
\midrule
\mname-D\subfrozen & 5\% & 0.001216 & 0.9635 & 0.9744 & 0.9729 \\
\mname-P\subfrozen & 5\% & 0.001204 & 0.9623 & 0.9717 & 0.9699 \\
\mname-D\subsmall  & 5\% & 0.005360 & 0.8387 & 0.7965 & 0.7952 \\
\mname-P\subsmall  & 5\% & 0.006917 & 0.8150 & 0.7531 & 0.7495 \\
MONet              & 5\% & 0.000544 & 0.9761 & 0.9752 & \textbf{0.9859} \\
MONet\subfrozen    & 5\% & \textbf{0.000324} & \textbf{0.9829} & \textbf{0.9753} & \textbf{0.9859} \\
\midrule
\mname-D\subfrozen & 1\% & \textbf{0.004970} & \textbf{0.8945} & \textbf{0.9351} & \textbf{0.9334} \\
\mname-P\subfrozen & 1\% & 0.007167 & 0.8444 & 0.8608 & 0.8591 \\
\mname-D\subsmall  & 1\% & 0.008119 & 0.7910 & 0.7094 & 0.7118 \\
\mname-P\subsmall  & 1\% & 0.009051 & 0.7778 & 0.6832 & 0.6788 \\
MONet              & 1\% & 0.020961 & 0.7654 & 0.6390 & 0.7764 \\
MONet\subfrozen    & 1\% & 0.012958 & 0.8277 & 0.7640 & 0.8457 \\
\bottomrule
\end{tabular}
\end{table}

\begin{figure}[t]
\centering
    \begin{subfigure}{0.495\textwidth}
        \centering
        \includegraphics[width=\textwidth, trim={0 0 274px 0}, clip]{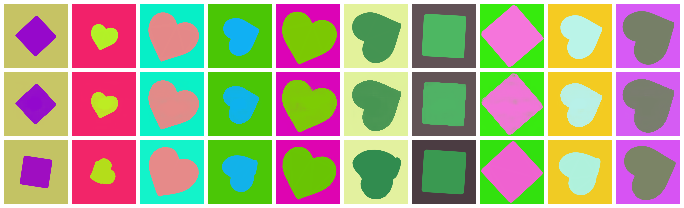}
        \caption{ Scene reconstructions produced by models trained on full training dataset. }
        \label{fig:problematic@1}
    \end{subfigure}
    \begin{subfigure}{0.495\textwidth}
        \centering
        \includegraphics[width=\textwidth, trim={0 0 274px 0}, clip]{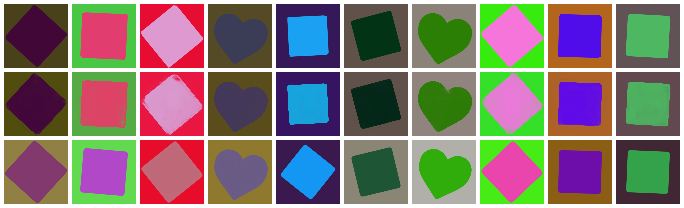}
        \caption{ Scene reconstructions produced by models trained on 5\% of the training data. }
        \label{fig:problematic@0.05}
    \end{subfigure}

    \begin{subfigure}{0.495\textwidth}
        \centering
        \includegraphics[width=\textwidth, trim={0 0 274px 0}, clip]{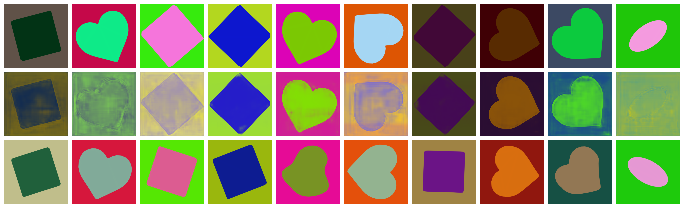}
        \caption{ Scene reconstructions produced by models trained on 1\% of the training data. }
        \label{fig:problematic@0.01}
    \end{subfigure}
    \caption{The reconstructions for 6 test-set examples that \mname fared worst on in terms of MSE, for models trained on 5\% (a) and 1\% (b) of the training set. Row-wise: the input image; the output of MONet\subfrozen; the output of \mname-P\subfrozen. }
    \label{fig:problematic_1}
\end{figure}

In Fig. \ref{fig:problematic_1}, we present the rendering of selected test-set examples produced by one of the \mname models (\mname-P\subfrozen) and compare it with one of the baselines (MONet\subfrozen). As the best renderings produced by all configurations are virtually indistinguishable from the input image, we focus on the worst cases, i.e. the 6 examples rendered with the largest MSE error by \mname-P\subfrozen. For the models trained on 5\% of data, the differences between the reconstructions produced by \mname and MONet can be traced back to their different operating principles: MONet is better at reproducing colors, but worse at modeling the shape of the objects. On the other hand, \mname can occasionally fail to predict the correct rotation of the object. For the models trained on 1\% of data, \mname can mangle the shape, while MONet may struggle with figure-ground separation, producing incorrect masks that blur the object and the background. It is important to note that the examples with the largest MSE error contain large objects, as pixel-wise metrics roughly correlate with object size.

\begin{figure}[t]
    \centering
    \begin{tikzpicture}
        \node at (0, 0) {\includegraphics[width=\textwidth]{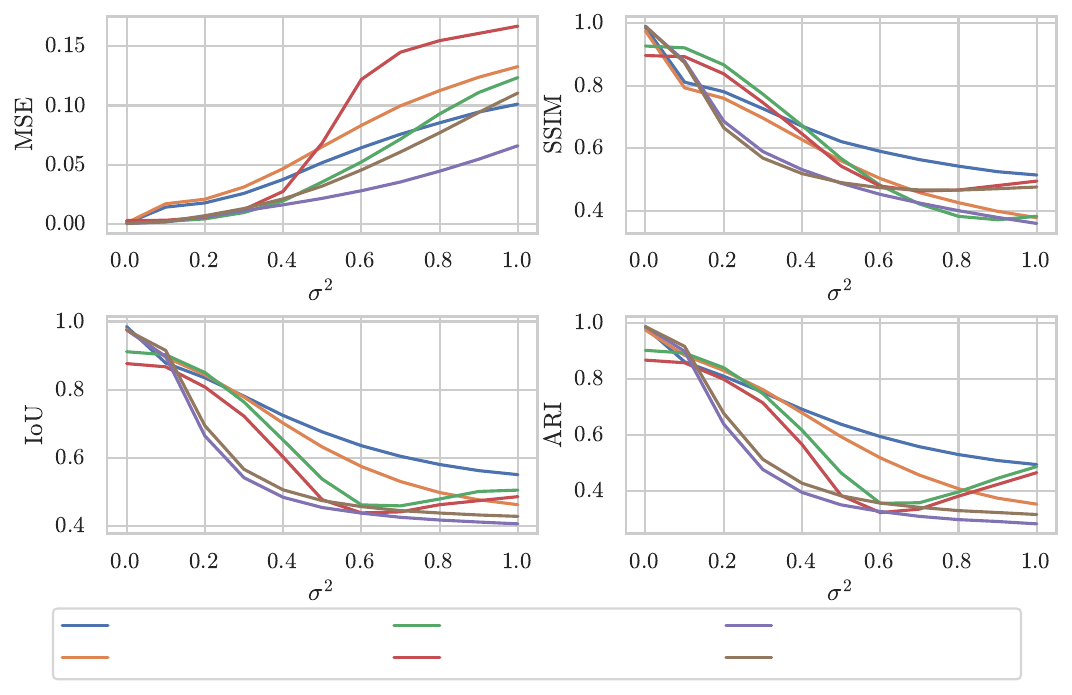}};
        \node[anchor=north west] at (-4.9, -2.95) {\relscale{0.75} \mname-D\subfrozen};
        \node[anchor=north west] at (-4.9, -3.35) {\relscale{0.75} \mname-P\subfrozen};
    
        \node[anchor=north west] at (-1.2, -2.95) {\relscale{0.75} \mname-D\subsmall};
        \node[anchor=north west] at (-1.2, -3.35) {\relscale{0.75} \mname-P\subsmall};

        \node[anchor=north west] at (2.6, -2.95) {\relscale{0.75} MONet};
        \node[anchor=north west] at (2.6, -3.35) {\relscale{0.75} MONet\subfrozen};
    \end{tikzpicture}
    \caption{ The impact of introducing noise to the test-set on the metrics. Noise was sampled from normal distribution with mean $0$ and standard deviation $\sigma^2$.}
    \label{fig:noise}
\end{figure}
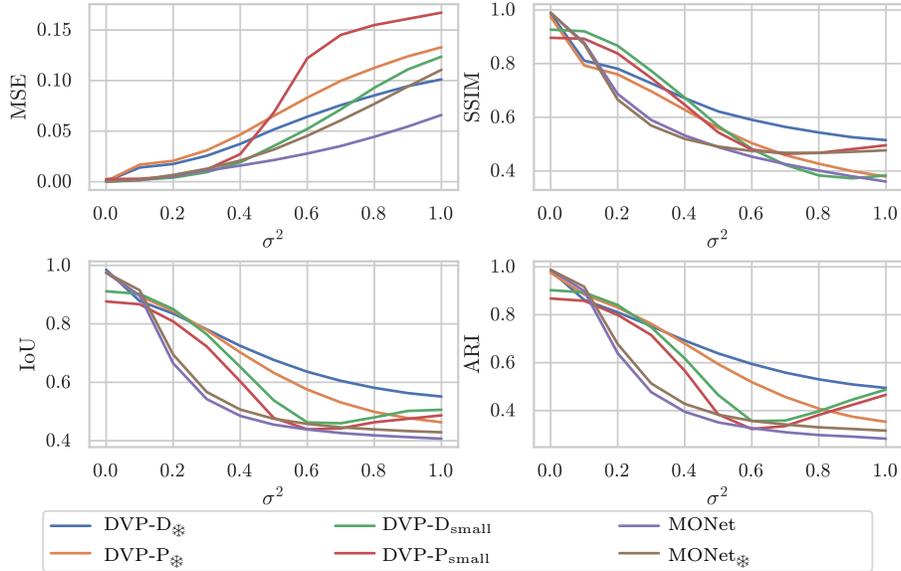

\vspace{1mm}\noindent\textbf{Robustness}. Figure \ref{fig:noise} presents how the metrics of the models trained on all data (100\%) degrade with the increasing standard deviation $\sigma^2$ of normally distributed white noise added to pixels of test-set images. While MONet exhibits the best robustness on MSE, \mname is better on the qualitative metrics (IoU, SSIM and ARI), degrading more gracefully. 

\begin{figure}[t]
\begin{subfigure}{\textwidth}
        \centering
        \includegraphics[width=\textwidth]{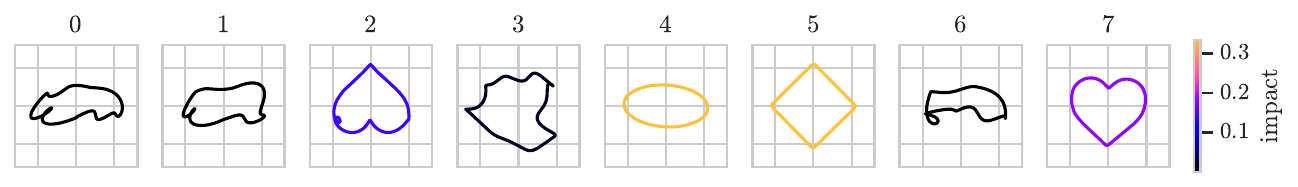}
        \label{fig:prototypes@1}
    \end{subfigure}\vspace{-6mm}
    \begin{subfigure}{\textwidth}
        \centering
        \includegraphics[width=\textwidth]{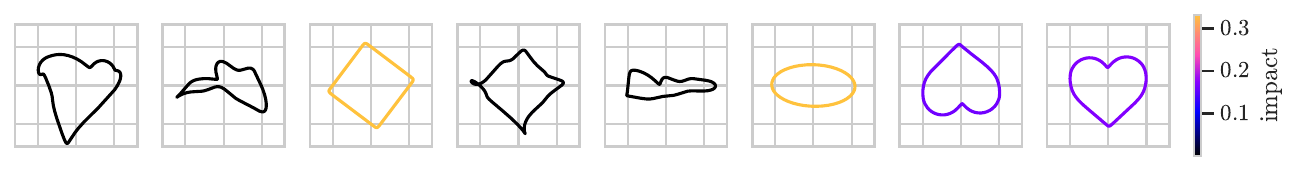}
        \label{fig:prototypes@0.05}
    \end{subfigure}\vspace{-6mm}
    \begin{subfigure}{\textwidth}
        \centering
        \includegraphics[width=\textwidth]{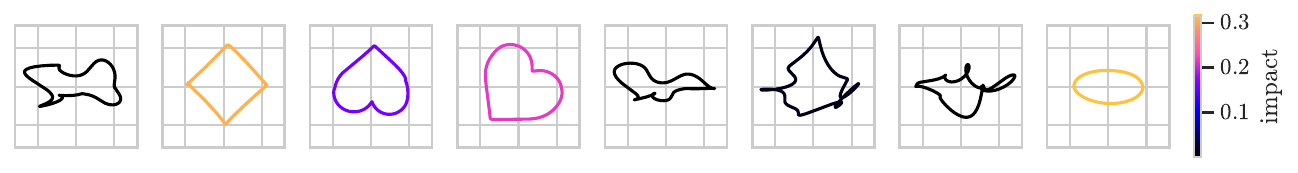}
        \label{fig:prototypes@0.01}
    \end{subfigure}\vspace{-5mm}
    \caption{ The prototypes learned by \mname-P\subfrozen trained on 100\% (top), 5\% (middle), and 1\% (bottom) of training data. Color represents the overall impact, i.e. the normalized sum of weights assigned to each prototype embedding by the Prototype function, estimated from the test-set. The order of prototypes is irrelevant.}
    \label{fig:prototypes}
\end{figure}

\vspace{1mm}\noindent\textbf{Explanatory capacity}. 
Figure \ref{fig:prototypes} shows the visual representation of the prototypes formed by \mname-P\subfrozen in its learnable 8-element embedding. The EFDs represent them as closed curves (Sec.\ \ref{app:efd}), which may occasionally coil (e.g. \#5 and \#6 at the bottom of Fig.\ \ref{fig:prototypes}). All presented models, including the one trained on just 1\% of data, learned prototypes that correctly capture shape categories. The remaining embedding slots contain random curves, used sparingly and contributing only marginally to the predicted shape, as evidenced by the normalized sum of embedding weights visualized in color. Models usually allocate a single embedding slot per category, except for hearts, for which they often form two prototypes. Given that these prototypes are rotated in opposite (or almost opposite) directions and used alternatively (notice lower weights), we posit that in the early stages of learning, the hearts' prototype is an equilateral triangle, and Rotation co-adapts to the 120\degree\xspace invariance of this shape by generating a limited range of rotation angles. Once the prototype shape becomes more accurate, that invariance is lost, and it is easier to form a second prototype than to re-learn Rotate. 

By assigning labels to the identified categories, we can use a \mname model as a classifier that points to the predicted category with the $\arg\max$ over the outputs of the MLP in the Prototype function. We determined that all \mname models presented in Table \ref{tab:accuracy}, when queried in this mode, achieve classification accuracy of 99.7\% or more when queried on the test set.

\begin{figure}[t]
    \centering
    \includegraphics[width=0.85\textwidth]{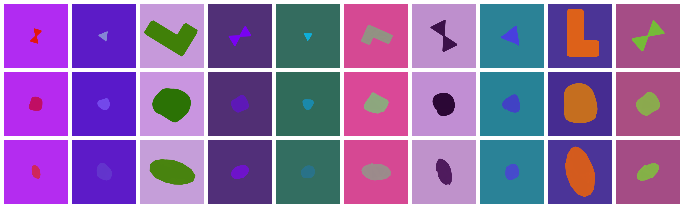}
    \caption{ Reconstructions for out-of-sample objects created by replacing shapes in the first 10 testing examples with hourglass, triangle, and L-shape. Row-wise: input scene; the output of \mname-D\subfrozen; the output of \mname-D\subsmall. }
    \label{fig:out-of-sample}
\end{figure}

\vspace{1mm}\noindent\textbf{Out-of-sample generalization}. To determine if the disentanglement of image formation aspects helps \mname to generalize well beyond the training distribution, we query selected variants of \mname and the baseline configurations on shapes from previously unseen categories: hourglass, triangle, L-shape. Table \ref{tab:out-of-sample} and Fig.\ \ref{fig:out-of-sample} summarize the quality of reconstruction. As expected, the metrics are worse than in Table \ref{tab:accuracy};  however, visual inspection of the reconstructed scenes reveals that \mname not only correctly models the background and the foreground colors, but also makes reasonably good predictions about object scale/size and orientation. Shape is the only aspect that is not modeled well enough. Interestingly, while \mname-D\subsmall substantially outperforms \mname-D\subfrozen on metrics, the latter is more faithful at reconstructing shape. Overall, the results confirm that \mname effectively disentangles the visual aspects also when faced with new types of objects. 


\begin{table}[t]
\centering
\caption{ Comparison of models on out-of-sample shape categories. The best results are highlighted in bold. }
\label{tab:out-of-sample}
\begin{tabular}{lcrrrrr}
\toprule
Method & MSE & SSIM & IoU & ARI & \\
\midrule
\mname-D\subfrozen & 0.009847 & 0.8031 & 0.4910 & 0.5858 \\
\mname-P\subfrozen & 0.013051 & 0.7810 & 0.4364 & 0.5297 \\
\mname-D\subsmall  & 0.006580 & 0.8368 & 0.5373 & 0.6170 \\
\mname-P\subsmall  & 0.006772 & 0.8303 & 0.5130 & 0.5922 \\
MONet              & \textbf{0.001357} & \textbf{0.9554} & 0.9026 & 0.9203 \\
MONet\subfrozen    & 0.002064 & 0.9482 & \textbf{0.9308} & \textbf{0.9359} \\
\bottomrule
\end{tabular}
\end{table}

\vspace{1mm}\noindent\textbf{Discussion}. 
While conventional DL models still maintain the upper hand when compared with \mname on pixel-wise metrics (Table \ref{tab:accuracy}), it is important to emphasize that the precise reconstruction of all minutiae of the image content is not the primary goal here. Reconstruction error serves here only as the guidance for the learning process, and in most use cases robust information about scene structure and composition will be of more value than attention to detail. Moreover, having a correctly inferred scene structure significantly facilitates further processing, like precise segmentation of individual objects with conventional techniques. 

One of the key advantages of \mname is \emph{transparency} and \emph{explainability}. For the sake of \emph{global explanation}, each component of the model is by construction endowed with an a priori known interpretation. For \emph{local explanation}, the outputs produced by \mname components in response to a concrete image can be inspected and interpreted, as evidenced above by our analysis of the learned prototypes. \mname produces an \emph{`evidence-based', compositional interpretation of the scene} that can be verified and reasoned about. 

Thanks to task decomposition provided by DSL programs, \mname can disentangle image formation aspects using a simple pixel-wise loss function, rather than resorting to more sophisticated means. This disentanglement addresses the combinatorial explosion of the number of interactions of shape, size, orientation, and appearance. The DSL program informs the model about the way they interact with each other, and so facilitates learning and generalization, without any need for data labeling, tagging, or other forms of supervision. In particular,  even though we endow each DSL function with a specific semantic (e.g. that Rotation controls the orientation), we do not train them via supervision, with concrete output targets --- the guidance they receive in training originates in the \emph{interactions} with other DSL functions they are composed with.  

Compared to conventional disentangling autoencoders (like the Variational Autoencoder, VAE \cite{Kingma_Welling_2013}), the disentanglement in \mname is arguably not \emph{entirely} emergent, as it is guided by a \emph{manually designed} DSL program. In this sense, \mname offers `explanation by design'. Notice, however, that explanation always requires pre-existing domain knowledge. 
If, for instance, one strives to determine whether a DL model has learned the concept of object rotation, that concept must be first known \emph{to him/her}. In other words, one can equip the model with snippets of domain knowledge in advance or look for them in the model only once it has been trained. Our approach follows the first route, offering both explanation and efficient learning.


\section{Conclusions and future work}\label{sec:conclusions}

This study demonstrated our preliminary attempt at developing a compositional and versatile DSL framework for well-founded image interpretation. Our long-term goal is the structural decomposition of complex scenes (prospectively 3D), involving multiple composite objects, alternative object representations, and other aspects of image formation (e.g. texture, object symmetry). In particular, parsing more complex scenes than those considered here will require extending the DSL to allow analysis of multiple objects while resolving the ambiguities that may originate in, among others, occlusion. In general, one may expect different types of scenes to be more amenable to interpretation by different DSL programs. For ambiguous scenes (e.g. due to occlusion), there might be multiple alternative interpretations. For these reasons, we intend to equip \mname with a generative program synthesis module that will produce alternative scene parsing DSL programs in a trial-and-error fashion, guided by feedback obtained from the confrontation of the produced rendering with the input image. 

\begin{credits}

\subsubsection{\ackname} 
This research was supported by TAILOR, a project funded by EU Horizon 2020 research and innovation program under GA No. 952215, by the statutory funds of Poznan University of Technology and the Polish Ministry of Education and Science grant no. 0311/SBAD/0726.
\end{credits}

\newpage
\begin{appendix}
\renewcommand{\thesection}{\appendixname \ \arabic{section}}

\section{Technical details of \mname}\label{app:model}

\subsection*{Architecture}

\vspace{1mm}\noindent\textbf{Perception module}. The perception module is composed of CNN used for feature extraction and a submodule used for mapping those features to latent vector $z$ (Fig.\ \ref{fig:diagram}). 

Both \mnamens\subsmall configurations employ a CNN1 architecture, which consists of repeating the following block: a convolutional layer with a kernel size of $3\times3$, a GELU\footnote{Gaussian Error Linear Unit.} activation function, an average pooling layer with a window size of $2\times2$, and a batch normalization layer. This block is repeated six times, while increasing the number of output channels: 64, 128, 256, 256, 512, 512. This is followed by a single convolutional layer with a kernel size of $1\times1$ (equivalent to a linear layer applied per pixel) to reduce the number of output channels to 256. The resulting tensor of size $1\times1\times256$ is flattened, forming a 256-dimensional $z$ vector.

\mnamens\subfrozen configurations use ConvNeXt-B as the feature extraction submodule. We use the pretrained instance of this network which was trained on the ImageNet-1K dataset. The feature extractor is frozen, which means its weights are not updated during the optimization process. The feature map produced by the extractor is extended with spatial positional encoding, flattened, and passed to the Transformer submodule, whose task is to transform the feature map into 256-dimensional $z$ vector. This approach is inspired by Detection Transformer (DETR) \cite{detr}; our Transformer submodule reuse DETR's hyperparameters.

\vspace{1mm}\noindent\textbf{Learnable functions}. All learnable functions of the DSL use 3-layer MLPs with a hidden layers' size of 256 and GELU activation function. The output layer is designed to match the needs of a given DLS function in terms of the number of units and activation function. For instance, the ObjectAppearance function produces 3-dimensional vectors in the range $[0, 1]$, therefore its MLP has 3 units in the last layer, each equipped with the sigmoid activation function.

\subsection*{Training}

Training \mname models in a generic way, i.e. starting from default random initialization of all parameters and using bare MSE as the loss function, leads on average to worse results than those reported in Table \ref{tab:accuracy}. To attain the reported level of accuracy, \mname needs additional guidance, particularly in its prototype-based variant \mname-P. While these aspects are usually not critical for progress in training and its convergence, we cover them in this section for completeness.  

\mname-D configurations require relatively little guidance. Initially, EFD shape contours produced by the model often appear 'jagged' and contain many intersections and loops. To address this issue, we add to the main MSE loss function an extra component that encourages the model to increase the amplitude of the first component and penalizes the subsequent low-frequency components (i.e., in the absence of MSE error, this component would in the limit cause the model to produce perfect circles). This form of regularization is applied with a weight of 0.001 while processing the first 30 kimg\footnote{1 kimg = 1024 images} in training.

\mname-P configurations require more supervision and hyperparameter tuning. We apply the following techniques:
\begin{itemize}
    \item The prototypes are initialized with random hexagons (even though technically speaking, the EFD order used in our configurations (8) is insufficient to precisely model all the corners of these polygons, resulting in rounded shapes).
    \item The prototypes are frozen for the first 30 kimg of training in case of all \mname-P\subfrozen models, and respectively 480 kimg, 240 kimg and 60 kimg in case of \mname-P\subsmall models trained on 100\%, 5\%, and 1\% of the training set. Without freezing, the prototypes tended to collapse to a local minimum (a circle), which made it difficult to learn how to rotate them. 
    \item We employed the load balancing loss \cite{switch_transformer} to encourage the P prototype-weighing MLP in the Prototype DSL function to choose the prototypes uniformly. The balancing loss is turned off after 120 kimg for \mname-P\subfrozen models, and respectively after 960 kimg, 480 kimg and 240 kimg in case of \mname-P\subsmall models trained on 100\%, 5\%, and 1\% of the training set.
    \item As illustrated in Fig.\ \ref{fig:prototypes}, the models occasionally reconstruct the shapes as mixtures of multiple prototypes. In order to address this issue, we apply \emph{distribution sharpening} to the distribution produced by the prototype-weighing MLP. The sharpening starts at the 5th epoch for \mname-P\subfrozen models, the 10th epoch for \mname-P\subsmall trained on 5\% and 1\% of the training set, and the 20th epoch for \mname-P\subsmall trained on the full training set.
    \item We apply gradient clipping by norm to the prototypes with the maximum norm of 0.01 in order to stabilize the training.
\end{itemize}

\section{Representing shapes with Elliptic Fourier Descriptors}\label{app:efd}

The elliptic Fourier Transform \cite{kuhl1982} is a method of encoding a closed contour with Fourier coefficients. The method can be viewed as an extension of the discrete-time Fourier Transform from the time domain to the spatial domain. We assume the contour to be encoded with the transform to be represented as a sequence of $K$ contour points $(x_p,y_p)$ such that $x_1 = x_K$ and $y_1 = y_K$. The elliptic Fourier transform of order $N$ is defined as:

\begin{equation}
\label{eqn:efd}
\begin{gathered}
A_n = \frac{T}{2 n^2 \pi^2} \sum_{p=1}^{K} \frac{\Delta x_p}{\Delta t_p} \left(\cos\frac{2n\pi t_p}{T} - \cos\frac{2n\pi t_{p-1}}{T} \right)  \\
B_n = \frac{T}{2 n^2 \pi^2} \sum_{p=1}^{K} \frac{\Delta x_p}{\Delta t_p} \left(\sin\frac{2n\pi t_p}{T} - \sin\frac{2n\pi t_{p-1}}{T} \right)  \\
C_n = \frac{T}{2 n^2 \pi^2} \sum_{p=1}^{K} \frac{\Delta y_p}{\Delta t_p} \left(\cos\frac{2n\pi t_p}{T} - \cos\frac{2n\pi t_{p-1}}{T} \right) \\
D_n = \frac{T}{2 n^2 \pi^2} \sum_{p=1}^{K} \frac{\Delta y_p}{\Delta t_p} \left(\sin\frac{2n\pi t_p}{T} - \sin\frac{2n\pi t_{p-1}}{T} \right) \\
\end{gathered}
\text{ for } n = 1,2,...,N
\end{equation}

\noindent where 
$\Delta x_p \equiv x_p - x_{p-1}$, $\Delta y_p \equiv y_p - y_{p-1}$, $\Delta t_p = \sqrt{\Delta x_p^2 + \Delta y_p^2}$ and $T = \sum_{p=1}^{K}\Delta t_p$. 
\vspace{3mm}


The Elliptic Fourier Descriptors (EFD) are the coefficients $A_n$, $B_n$, $C_n$, and $D_n$. They are translation invariant by design and can be further normalized to be invariant w.r.t. rotation and scale.

The original contour can be reconstructed using the inverse transform given by the following equations:

\begin{equation}
\begin{gathered}
x_p = \sum_{n=1}^{N} \left(\ A_n \cos\frac{2n\pi t_p}{T} + B_n \sin\frac{2n\pi t_p}{T} \right) \\
y_p = \sum_{n=1}^{N} \left(\ C_n \cos\frac{2n\pi t_p}{T} + D_n \sin\frac{2n\pi t_p}{T} \right) \\
\end{gathered}
\text{ for } p = 1,2,...,K
\end{equation}

\end{appendix}

\bibliographystyle{splncs04}
\bibliography{bibliography}

\end{document}